# Syllable Subword Tokens for Open Vocabulary Speech Recognition in Malayalam

**Kavya Manohar**[*]
College of Engineering
Trivandrum, India
sakhi.kavya@gmail.com

**A. R. Jayan**
Government Engineering
College, Thrissur, India
arjayan@gectcr.ac.in

**Rajeev Rajan**
Government Engineering
College, Trivandrum, India
rajeev@cet.ac.in

## Abstract

In a hybrid automatic speech recognition (ASR) system, a pronunciation lexicon (PL) and a language model (LM) are essential to correctly retrieve spoken word sequences. Being a morphologically complex language, the vocabulary of Malayalam is so huge and it is impossible to build a PL and an LM that cover all diverse word forms. Usage of subword tokens to build PL and LM, and combining them to form words after decoding, enables the recovery of many out of vocabulary words. In this work we investigate the impact of using syllables as subword tokens instead of words in Malayalam ASR, and evaluate the relative improvement in lexicon size, model memory requirement and word error rate.

## 1 Introduction

Malayalam belongs to the Dravidian family of languages with high morphological complexity (Manohar et al., 2020). Productive word formation in Malayalam by agglutination, inflection, and compounding leads to very long words with phonetic and orthographic changes at morpheme boundaries. This creates a large number of low frequency words and it is practically impossible to build a pronunciation lexicon that covers all complex wordforms.

A hybrid automatic speech recognition (ASR) decoder is built using an acoustic model, a language model (LM) and a pronunciation lexicon (PL). The acoustic model is a mapping between the acoustic features and the phonemes of the language (Georgescu et al., 2021). The LM is a learnt representation of word sequence probabilities. The PL is a dictionary where the pronunciation of each word or subword is described as a sequence of phonemes. These are composed into a weighted finite state transducer in a typical hybrid ASR decoder (Povey et al., 2011).

[*]Also affiliated with APJ Abdul Kalam Technological University, Kerala, India

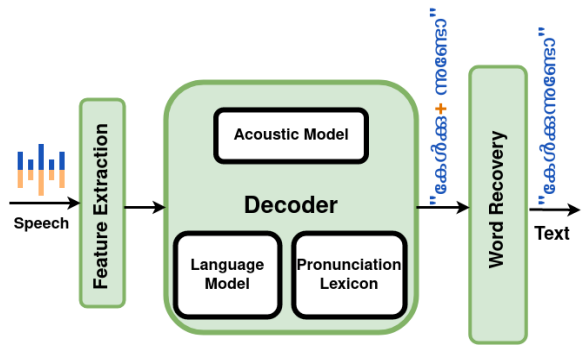

Figure 1: An open vocabulary hybrid ASR system, with subword based LM and PL.

Words not covered in the LM and the PL are called the out of vocabulary (OOV) words and they can not be recovered by the ASR decoder (Braun et al., 2021; Smit et al., 2021). However the use of subword tokens in an ASR for morphologically complex languages can recover a portion of OOV words by combining subword tokens to words. Figure 1 illustrates a hybrid open vocabulary ASR system. Special marker symbol '+' at subword boundaries enables the recovery of words.

Subword tokenization is carried out either through linguistically motivated rule based approaches or language independent data-driven approaches (Smit et al., 2021). However, there is no single algorithm that works fine for all languages. Even though the usage of subword tokenization for open vocabulary ASR has been thoroughly investigated (Hirsimäki et al., 2006; Choueiter et al., 2006; Wang et al., 2020; Zhou et al., 2021), there has not been much exploration in this regard in Malayalam language.

**Algorithm 1** Syllable Tokenization Algorithm
---
1: **procedure** SYLLABLE BOUNDARY TAGGING
2:   c_v ← consonant + virama
3:   Type 1 ← <BoW> + vowel+[anuswara, visarga, chillu] ?             ▷ ? indicates optionality
4:   Type 2 ← consonant + vowelsign ? + [anuswara, visarga, chillu]?
5:   Type 3 ← c_v * + consonant                                        ▷ * indicates one or more occurence
6:   Type 4 ← c_v ? + consonant + ◌്? + virama + <EoW>
7:   syllable ← [Type 1, Type 2, Type 3, Type 4]                        ▷ Defines a syllable
8:   `SyllableBoundaryTagger`: <BoS>+ syllable + <EoS> ← syllable
9: **end procedure**
---

## 2 Related Works

Morpheme based subword tokenization has been proposed for ASR in many morphologically complex languages including Finnish, Arabic and Swedish (Choueiter et al., 2006; Smit et al., 2021). Syllable like units called vowel segments have been proposed to improve the ASR performance of Sanskrit, which is an inflectional language (Adiga et al., 2021). Data driven methods of tokenization using byte pair encoding (BPE) and Morfessor has been employed in the development of bilingual Hindi-Marathi ASR for improved performance and reduced complexity (Yadavalli et al., 2022). The sole work on the usage subword tokens for Malayalam ASR (Manghat et al., 2022) applies the linguistic information on a data-driven method to improve the word error rate (WER).

In the current work, we investigate the improvement that can be brought in by the linguistically motivated syllable subword tokens to address the issue of OOV recovery in Malayalam ASR. We evaluate the syllable subword ASR in terms of the WER, the lexicon size and the model memory requirement and compare it with the conventional word based PL and LM. This work is planned to be extended to analyse the impact of other data-driven methods for subword tokenization, in future.

## 3 Tokenization Algorithm

The characters in Malayalam script can be classified as: (i) vowels, (ii) vowel signs, (iii) consonants, (iv) special consonants (*anuswara*, *visarga* and *chillu*) and (v) the multi-functional character *virama*. A conjunct in Malayalam is a sequence of consonants separated by a *virama* in between. The writing system of Malayalam is alphasyllabary in nature (Bright, 1999). It means each standalone pronunciation unit is a syllable. If words are randomly split during tokenization, as in **SOPHIA** /soʊfiə/ being tokenized as **SOP** and **HIA**, the pronunciation can not be segmented in a valid way. Syllable tokens being valid pronunciation units, they can be described as a sequence of phonemes in the PL.

A syllable in Malayalam can be a consonant or a conjunct, followed by an optional vowel sign. A standalone vowel is also a syllable, that occur only at word beginnings. Whenever a special consonant appears, it becomes the syllable ending consonant (Nair, 2016). These linguistic rules for syllable tokenization has been computationally implemented as in Algorithm 1, by Manohar et al. (2022) and made available as part of the Mlphon Python library[1].

## 4 Datasets

We use the publicly available open licensed Malayalam read speech datasets in our experiments. Every speech recording is associated with a textual transcript in the Malayalam script. As shown in Table 1, we divide the available speech into train and test, ensuring that speakers and speech transcripts are not overlapped. The train datasets are combined to get 1125 minutes (≈ 19 hours) of speech for acoustic model training. T1, T2 and T3 are the datasets used for testing. Except T3, all datasets are studio recorded read speech of formal sentences belonging to the same domain. T3 is mostly conversational sentences, recorded by volunteers in natural home environments, making it an out-of-domain test set.

To create the LM, we use the sentences from the speech transcripts and combine it with the curated collection of text corpus published by SMC (Computing, 2020a) that amounts to 205k unique sentences. From this, every sentence that appears in our test speech dataset is explicitly removed to prevent overfitting.

---
[1] https://pypi.org/project/mlphon/

Table 1: Details of Speech datasets used in our experiments.

| Name | Corpus | #Speakers | #Utterances | Duration (minutes) | Environment |
|---|---|---|---|---|---|
| Train 1 | (Baby et al., 2016) | 2 | 8601 | 838 | Studio |
| Train 2 | (He et al., 2020) | 37 | 3346 | 287 | Studio |
| T1 | (Prahallad et al., 2012) | 1 | 1000 | 98 | Studio |
| T2 | (He et al., 2020) | 7 | 679 | 48 | Studio |
| T3 | (Computing, 2020b) | 75 | 1541 | 98 | Natural, Noisy |

## 5 Methodology

To develop a hybrid ASR system, we need to build an acoustic model, an LM and a PL. The acoustic model is set as a common component in both word and syllable token based ASR. The LM is a statistical ngram model of words or syllables. To study the impacts of lexicon size we create word and syllable token based PL of different sizes. Each of these components is explained in the following subsections.

### 5.1 Acoustic Model

The acoustic model is trained using time delay neural networks (TDNNs) with Kaldi toolkit (Povey et al., 2011). Acoustic features used in TDNN training are: (i) 40-dimensional high-resolution MFCCs extracted from frames of 25 ms length and 10 ms shift and (ii) 100-dimensional i-vectors computed from chunks of 150 consecutive frames (Saon et al., 2013). Three consecutive MFCC vectors and the i-vector corresponding to a chunk are concatenated, resulting in a 220-dimensional feature vector for a frame (Georgescu et al., 2021). This acoustic model is trained on a single NVIDIA Tesla T4 GPU.

### 5.2 Language Models

A statistical view of how words are combined to form valid sentences is provided by the ngram model. Word sequence probabilities could be computed by analysing a large volume of text. In a 2-gram, a history of one previous word is required. We build ngram language models of orders n=2, 3 and 4 on the text corpus described in section 4 using SRILM toolkit (Stolcke, 2002).

Building LM using word tokens is straightforward, as *space* is considered as the default delimiter between words. However to build LM using syllable tokens instead of words, we need to syllabify the text corpus. Using Mlphon Python library, the text corpus is tokenized to syllables (Manohar et al., 2022). In order to identify syllables that occur at word medial positions, we have used '+' as a marker symbol.

Table 2: Samples of text from LM training corpus

| Word | അവൻ വള ഇടുകയില്ല |
| | /aʋan ʋaɭa iʈukajilla/ |
| Syllable | അ+ വൻ വ+ ള ഇ+ ടു+ ക+ യി+ ല്ല |
| | /a+ ʋan ʋa+ ɭa i+ ʈu+ ka+ ji+ lla/ |

In this approach, reconstruction of words is straightforward, as the marker indicates the positions for joining the following syllable. In the syllabified text, *space* is the delimiter between syllable tokens. Excerpts from the text copora used for training word and syllable token based LM are shown in Table 2.

### 5.3 Pronunciation Lexicons

Table 3: An excerpt from word PL and corresponding syllable PL.

| Word PL | | Syllable PL | |
|---|---|---|---|
| അവൻ | a ʋ a n | അ+ | a |
| വള | ʋ a ɭ a | വൻ | ʋ a n |
| ഇടുകയില്ല | i ʈ u k a j i l l a | വ+ | ʋ a |
| | | ള | ɭ a |
| | | ഇ+ | i |
| | | ടു+ | ʈ u |
| | | ക+ | k a |
| | | യി+ | j i |
| | | ല്ല | l l a |

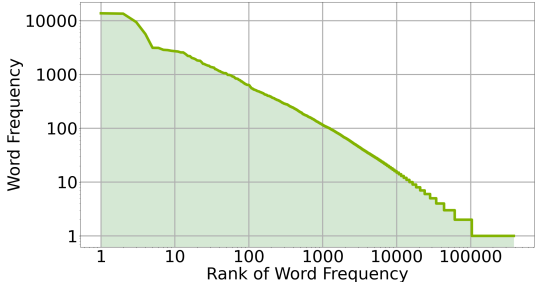

Figure 2: Logarithmic plot of word rank versus word frequency in the text corpus.

Sample entries in word PL and corresponding syllable PL are described in Table 3. To begin with, we create a word based PL that contains all unique words in the train audio transcripts which amounts to 25604 entries. This first lexicon is referred to as $PL1_W$. To study the impact of lexicon size on OOV rate and corresponding changes in WER, we expand $PL1_W$. New words are added to the lexicon based on their frequencies in the LM training corpus. When words in this corpus is ranked in the order of their frequencies, we obtain a word frequency profile as shown in Figure 2.

It can be seen that a huge portion of the corpus is covered by filling the PL with high frequency words. We add words with at least 5, 4, and 3 occurrences to $PL1_W$ to obtain the pronunciation lexicons $PL2_W$, $PL3_W$ and $PL4_W$ respectively. Subword lexicons $PLi_S$, with syllables as entries are derived from $PLi_W$, where $i = 1, 2, 3, 4$. The unique list of syllable tokens from every word PL is obtained to create the corresponding syllable PL. The number of entries in the syllable and word PL are presented in Table 4.

Table 4: The size of lexicons used in word and syllable based experiments

| Lexicon | Size | Lexicon | Size |
|---------|------|---------|------|
| $PL1_W$ | 25604 | $PL1_S$ | 3524 |
| $PL2_W$ | 53240 | $PL2_S$ | 5247 |
| $PL3_W$ | 62483 | $PL3_S$ | 5643 |
| $PL4_W$ | 79950 | $PL4_S$ | 6351 |

The syllable tokens corresponding to each word in $PLi_W$, is created with the marker symbol '+', as described in section 5.2. The pronunciation of word and syllable tokens in $PLi_W$ and $PLi_S$ are derived using Mlphon python library (Manohar et al., 2022).

## 6 Experimental Results

Combining the common acoustic model with the word LM, we build four different word based ASR by choosing one of $PLi_W$. Percentage of OOV words in different test datasets decreases with increase in the vocabulary size, as expected and is illustrated in Figure 3. Based on this, T1 can be considered as a low OOV test dataset, T2 a medium OOV test dataset and T3 a large OOV test dataset.

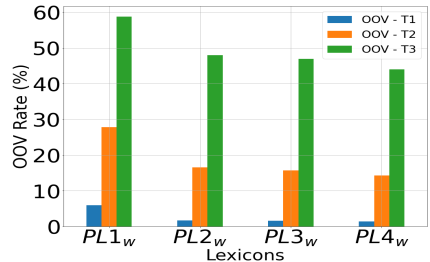

Figure 3: Lexicon size and OOV rate of test datasets

We repeat the above experiments with the LM training corpus and lexicons in syllabified form. Lexicons with syllables as entries are significantly smaller than word based lexicons, as indicated in Table 4 and are able to decode speech with improved WER on test datasets with medium to large OOV word rate. WER is computed by equation (1), based on the number of words inserted (I), deleted (D) and substituted (S) in the predicted speech when compared to the number of words (N) in ground truth transcript.

$$WER = \frac{I + D + S}{N} \qquad (1)$$

We report the WER on different test datasets in Figure 4. On the test set T1, where OOV rates are very low (less than 6%), word based model perform well irrespective of ngram orders, the best being 9.8%, while the best WER given by syllable models on T1 is only 12%. It shows syllable tokens are not advantageous in terms of WER in low OOV scenarios. The WER is generally high as expected on the out of domain test set T3, where almost half the words are OOV and the recording environment is drastically different from the train datasets.

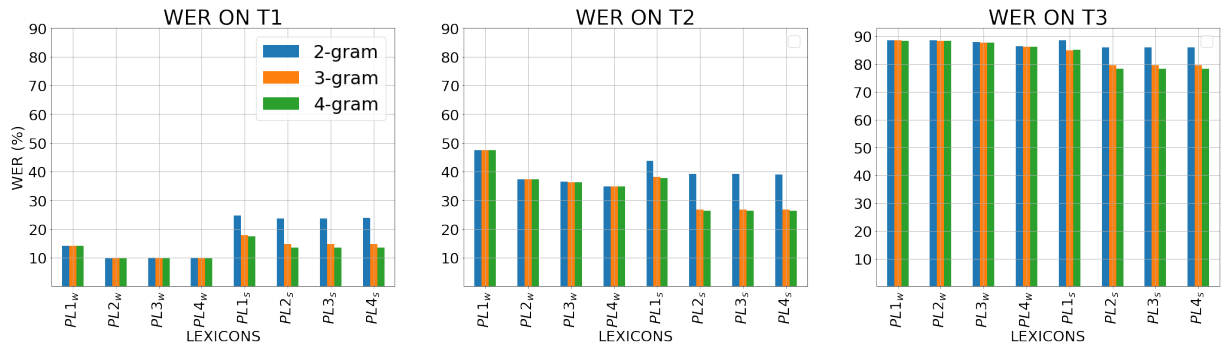

Figure 4: WER on different test datasets

Comparing the best WER, syllable based lexicons shows an improvement by 10% on T2 and by 7% on T3 than the corresponding word models. Since the previously published work on subword ASR for Malayalam (Manghat et al., 2022), was tested on a private dataset, the comparison of results is not meaningful and hence not attempted.

**Ngram order and WER**

For the word PL, increasing the ngram order imparts only nominal improvement in WER. This could be attributed to the sparse distribution of words due to the morphological complexity of Malayalam. The WER of the syllable PL does not show an improvement than the word PL for ngram order of 2. But the WER on syllable PL drastically reduces by 12% on T2 and by 6% on T3, when ngram order is increased from 2 to 3 and then it stabilizes. The mean word length in our test datasets is 3.2 syllables, providing the cause for the greatest improvement at this ngram transition.

**Ngram order and Model Size**

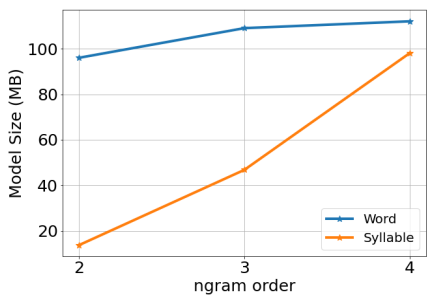

Figure 5: Model Size for word and syllable ASR.

To study the the model memory requirement, we compute the size of weighted FST graph (*HCLG.fst*) used for decoding. The model sizes corresponding to the largest word and syllable lexicons $PL4_W$ and $PL4_S$, where the WER are the best, are presented in Figure 5.

The memory requirement is generally high for word based models and it increases with the ngram order. The syllable tokens with much lower memory requirement at smaller ngram orders, show a rapid rise in model size with the increase in ngram order. There is a trade-off between the model size and the WER, while choosing the ngram order. For the ngram order of 3, ASR with syllable tokens having half the model size perform better in WER by 6% than the best word based model, as illustrated in Figures 4 and 5.

**Lexicon Size and WER**

There is a substantial WER improvement, when switching from $PL1_W$ to $PL2_W$ and $PL1_S$ to $PL2_S$, where the reduction in OOVs is the largest. Improvement in WER with subsequent lexicon expansions is nominal, as the added entries in the lexicons are low frequency words.

## 7 Conclusions

The comprehensive evaluation of syllables as subword tokens for building an open vocabulary hybrid ASR model is a pioneer attempt of its kind in Malayalam language. The proposed syllable based LM and PL in Malayalam demonstrate remarkable improvement in WER on medium and large OOV test sets, by 10% and 7% respectively . If the test datasets are free from OOV words, word based models outperform syllable models. Furthermore, syllable models with about half the model size have better WER than the corresponding word based ones, proving the effectiveness of syllable token based subword modelling on morphologically complex language like Malayalam. The optimal choice of ngram order based on the trade-off between model size and WER, depends on the subword tokenization technique. This study opens scope for investigating the impacts of other subword tokenization methods for Malayalam ASR.